\documentclass[11pt,a4paper]{article}
\usepackage[hyperref]{aacl-ijcnlp2020}
\usepackage{times}
\usepackage{latexsym}

\usepackage{microtype}
\aclfinalcopy 
\def\aclpaperid{300} 
\usepackage{amsmath}
\usepackage{amsfonts}
\usepackage{mathtools}
\usepackage{bm}
\usepackage{graphicx}
\usepackage{comment}
\usepackage{booktabs}
\usepackage{multirow}
\usepackage{subfig}
\usepackage[acronym]{glossaries}
\usepackage{enumitem}
\setlist{nosep}
\usepackage{array}
\newcolumntype{H}{>{\setbox0=\hbox\bgroup}c<{\egroup}@{}}
\DeclarePairedDelimiter\abs{\lvert}{\rvert}%
\DeclarePairedDelimiter\norm{\lVert}{\rVert}%
\makeatletter
\let\oldabs\abs
\def\abs{\@ifstar{\oldabs}{\oldabs*}}
\let\oldnorm\norm
\def\norm{\@ifstar{\oldnorm}{\oldnorm*}}
\makeatother
\glsdisablehyper
\newacronym{ic}{IC}{image captioning}
\newacronym{ric}{RIC}{relation-aware image captioning}
\newacronym{gnn}{GNN}{graph neural network}
\newacronym{gat}{GAT}{graph attention network}
\newacronym{cgat}{C-GAT}{conditional graph attention}
\newacronym{fa}{FA}{flat attention}
\newacronym{ha}{HA}{hierarchical attention}
\newacronym{hasg}{HA-SG}{hierarchical attention with scene graph first}
\newacronym{haim}{HA-IM}{hierarchical attention with objects detected first}
\newacronym{butd}{BUTD}{bottom-up top-down}
\newacronym{kmsl}{KMSL}{know more, say less}
\newacronym{nbt}{NBT}{neural baby talk}
\newacronym{sgg}{SGG}{scene graph generation}
\newacronym{imp}{IMP}{iterative message passing}
\newacronym{rpn}{RPN}{region proposal network}
\newacronym{relpn}{RelPN}{relation proposal network}
\title{Are scene graphs good enough to improve Image Captioning?}

\author{Victor Milewski$^1$ \ \ Marie-Francine Moens$^1$ \ \ Iacer Calixto$^{2,3}$\\\\
$^1$KU Leuven \ \
$^2$New York University \ \ 
$^3$ILLC, University of Amsterdam\\
\texttt{\{victor.milewski,sien.moens\}@cs.kuleuven.be} \\
\texttt{iacer.calixto@nyu.edu}
}

\date{June 26, 2020}
\begin{document}
\maketitle
\begin{abstract}
Many top-performing image captioning models rely solely on object features computed with an object detection model to generate image descriptions.
However, recent studies propose to directly use scene graphs to introduce information about object relations into captioning, hoping to better describe interactions between objects.
In this work, we thoroughly investigate the use of scene graphs in image captioning.
We empirically study whether using additional scene graph encoders can lead to better image descriptions and propose
a \textit{conditional} graph attention network (C-GAT),
where the image captioning decoder state is used to condition the graph updates. 
Finally, we determine to what extent noise in the predicted scene graphs influence caption quality.
Overall, we find no significant difference between models that use scene graph features and models that only use object detection features across different captioning metrics, which suggests that existing scene graph generation models are still too noisy to be useful in image captioning.
Moreover, although the quality of predicted scene graphs is very low in general, when using high quality scene graphs we obtain
gains of up to 3.3 CIDEr compared to a strong Bottom-Up Top-Down baseline.\footnote{We open source the codebase to reproduce all our experiments in \url{https://github.com/iacercalixto/butd-image-captioning}.}
\end{abstract}
\glsresetall
\glsunset{hasg}
\glsunset{haim}

\section{Introduction}
Scene understanding is a complex and intricate activity which humans perform effortlessly but that computational models still struggle with.
An important backbone of scene understanding is being able to detect objects and relations between objects in an image, and scene graphs~\citep{img-retrieval-scene-graph,anderson2016spice} are a closely related data structure that explicitly annotates an image with its objects and relations in context.
Scene graphs can be used to improve important visual tasks that require scene understanding, e.g. image indexing and search~\citep{img-retrieval-scene-graph} or scene construction and generation~\citep{Johnson_2017_clevr, image_generation_scene_graph}, and there is evidence that they can also be used to improve image captioning~\citep{yangetal2019autoencoding,know_more_say_less}.
However, the \textit{de facto} standard in top-performing image captioning models to date use strong object features only, e.g. obtained with a pretrained Faster R-CNN~\citep{faster-rcnn}, and no explicit relation information~\citep{bottomup_topdown,neural_baby_talk,Yuetal2019MultimodalTransformer}.

One possible explanation to this observation is that by using detected objects we already capture the more important information that characterises a scene, and that relation information is already \textit{implicitly} learned in such models.
Another explanation is that relations are simply not as important as we hypothesise and that we gain no valuable extra information by adding them.
In this work, we investigate these empirical observations in more detail and strive to answer the following research questions: 
(i) Can we improve image captioning by explicitly supervising a model with information about object relations?
(ii) How does the content of the captions improve when utilising scene graphs?
(iii) How does scene graph quality impact the quality of the captions?

The most recent best-performing image captioning models make use of the Transformer architecture~\citep{vaswani2017attention,Li_2019_ICCV, transformer_cap}.
However, in this paper we build upon the influential Bottom-Up Top-Down architecture \citep{bottomup_topdown} which uses LSTMs, and since we want to measure to what extent scene graphs are helpful or not, we remove any ``extras'' to make model comparison easier, e.g. reinforcement learning step after cross-entropy training, ensembling at inference time, etc.

\begin{figure*}[t!]
    \centering
    \includegraphics[width=.9\linewidth]{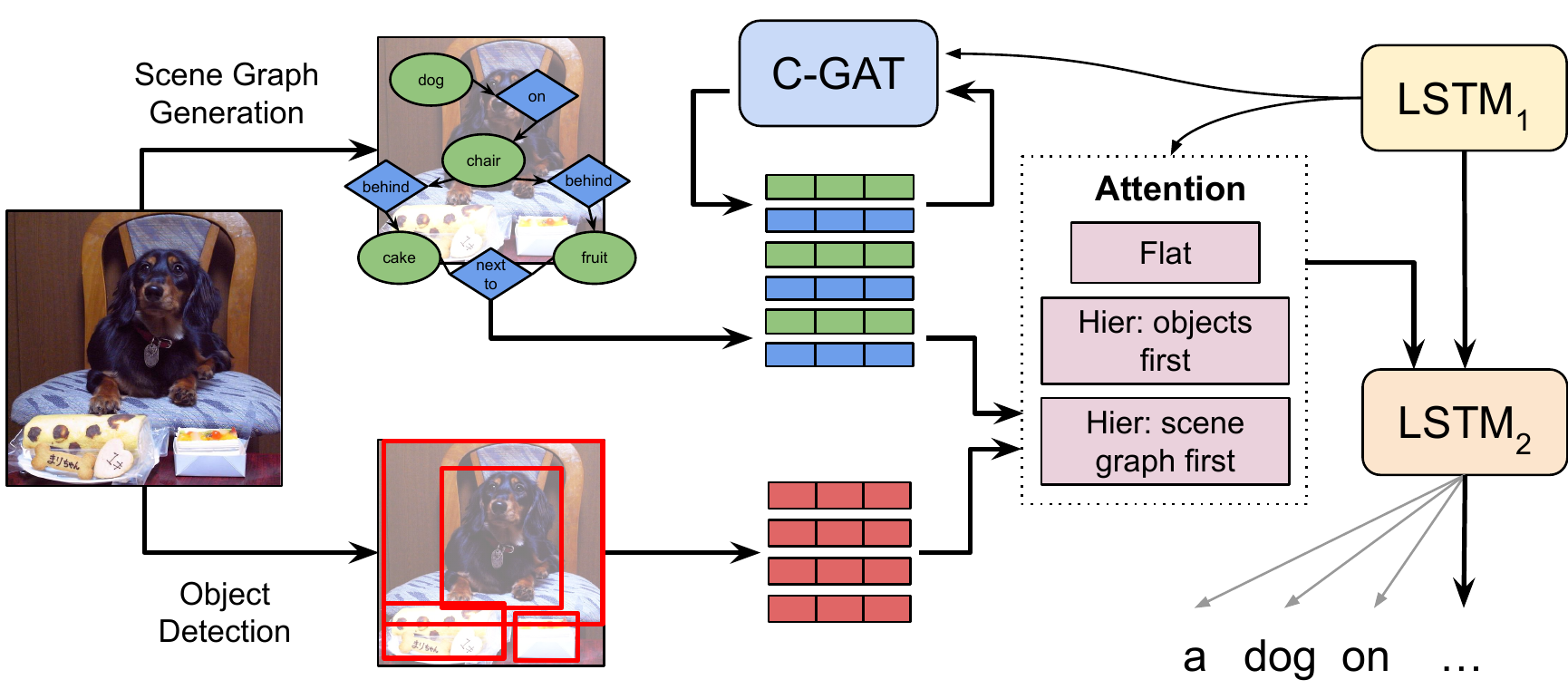}
    \caption{We use object features from an object detection model and scene graph features from a scene graph generation model in image captioning. We use \acrfull{cgat} to encode scene graph features, and flat vs. hierarchical attention mechanisms are used to incorporate both feature sets into an LSTM decoder.}
    \label{fig:cgat_model}
\end{figure*}

\Gls{sgg} is the task where given an image a model predicts a graph with its objects and their relations.
We use a pretrained \gls{sgg} model~\citep{scene-graph-iterative} to obtain and inject explicit relation information into image captioning, and investigate different image captioning model architectures that incorporate object and relation features, similarly to~\citet{know_more_say_less,wang-etal-2019-role}.
We propose an extension to graph attention networks \citep{graph_attention} which we call \gls{cgat}, where we condition the updates of the scene graph features on the current image captioning decoder state.
Finally, we conduct an in-depth analysis of the captions produced by different models and determine if scene graphs actually improve the content of the captions.
Our approach is illustrated in Figure~\ref{fig:cgat_model}.

Our main contributions are:
\begin{itemize}
    \item We investigate different graph-based architectures to fuse object and relation information derived from scene graph generation models in the context of image captioning. 
    \item We introduce conditional graph attention networks to condition scene graph updates on the current state of an image captioning decoder and find that it leads to improvements of up to 0.8 CIDEr.
    \item We compare the quality of the generated scene graphs and the quality of the corresponding captions and find that by using high quality scene graphs we can improve captions quality by up to 3.3 CIDEr.
    \item We systematically analyse captions generated by standard image captioning models and by models with access to scene graphs using SPICE scores for objects and relations~\citep{anderson2016spice} and find that when using scene graphs there is an increase of 0.4 F1 for relations and decrease of 0.1 F1 for objects.
\end{itemize}

\section{Background}\label{sec:background}

\subsection{Object Detection}\label{sec:object_detection}
Object detection is a task where given an input image the goal is to locate and label all its objects. 
The {\bf Faster R-CNN}, which builds on the R-CNN and Fast R-CNN~\citep{rcnn,fast_rcnn}, is a widely adopted model proposed for object detection~\citep{faster-rcnn}.
It uses a pretrained convolutional neural network (CNN) as a backbone to extract feature maps for an input image.
A \gls{rpn} uses these feature maps to propose a set of regions with a high likelihood of containing an object. 
For each region, a feature vector is generated using the feature map, which is then passed to an object classification layer. 
In our experiments, we use the Faster R-CNN with a ResNet-101 backbone~\citep{he2016resnet}.

\subsection{Graph Neural Networks}\label{sec:gnn}
Graph neural networks~\citep[\acrshortpl{gnn};][]{battaglia2018relational}\ifglsused{gnn}{}{\glsunset{gnn}} are neural architectures designed to operate on arbitrarily structured graphs  $\mathcal{G}=(\mathcal{V},\mathcal{E})$, where $\mathcal{V}$ and $\mathcal{E}$ are the set of vertices and edges in $\mathcal{G}$, respectively.
In \glspl{gnn}, representations for a vertex $v\in\mathcal{V}$ are computed by using information from neighbouring vertices $\mathcal{N}(v)$ which are defined to include all vertices connected through an edge.
In this work, we use a neighbourhood $\mathcal{N}(v_1)$ that contains vertices $v_2$ connected through \textit{incoming} edges, i.e. $v_2 \to v_1 \in \mathcal{E}$.
%
%
%

\paragraph{Graph Attention Networks}
\Acrlongpl{gat}~\citep[\acrshortpl{gat};][]{graph_attention}\ifglsused{gat}{}{\glsunset{gat}} combine features from neighbour vertices $\mathcal{N}(v_1)$ through an attention mechanism~\citep{bahdanau2014neural} to generate representations for vertex $v_1$.
Vertex $v_1$'s state $\bm{v}_1^{t-1}$ at time step $t-1$ is used as the query to soft-select the information from neighbours relevant to its updated state $\bm{v}_1^{t}$.

\subsection{Scene Graphs}\label{sec:scene_graphs}
Scene graphs consist of a data structure devised to annotate an image with its objects and the existing relations between objects and were first introduced for image retrieval~\citep{img-retrieval-scene-graph}.

We consider scene graphs $\mathcal{G}$ for an image with two types of \textit{vertices}: objects and relations.\footnote{Attributes are also originally present in scene graphs as vertices, but we do not use them.}
Object vertices describe the different objects in the image, and relation vertices describe how different objects interact with each other.
This gives us the following rules for edges $\mathcal{E}$: \textit{(i)} All existing edges are between an object vertex and a relation vertex;
\textit{(ii)} If an object \texttt{o$_1$} is connected to another object \texttt{o$_2$} via a relation vertex \texttt{r$_3$}, then vertex \texttt{r$_3$} has \textit{only} two connected edges: one incoming from \texttt{o$_1$} and one outgoing to \texttt{o$_2$}.
Finally, object (relation) vertices are also associated to an object (relation) label.

\paragraph{Scene Graph Generation}
\Acrfull{sgg} was introduced by~\citet{scene-graph-iterative} and has since received growing attention~\citep{neuralmotifs,li2018factorizable_net,graph-rcnn,knyazev2020graph-novel-composition}.
One can compare it to object detection (Section~\ref{sec:object_detection}), where instead of only predicting objects a model must additionally predict which objects have relations and what are these relations.
This similarity makes it natural that \gls{sgg} models build on object detection architectures.
Most \gls{sgg} models use a pretrained Faster R-CNN or similar architecture to predict objects and have an additional component to predict relations for pairs of objects.
In addition to the original object loss components in the Faster R-CNN, they include a mechanism to update object feature representations using neighbourhood information, and a component to predict relations and their label.

\paragraph{Iterative Message Passing}
The Iterative Message Passing \gls{sgg} model~\citep{scene-graph-iterative}
keeps two sets of states, i.e. for object vertices and relation vertices. The object vertices are initialised directly from Faster R-CNN features, while a relation vertex is computed by the box union of each of its two objects boxes, which is encoded with the Faster R-CNN to obtain a relation vector. 

Hidden states in each set are updated using an attention mechanism over neighbour vertices, i.e. objects are informed by all connected relation vertices, and relations are informed by the two objects it links. 
Since there are two sets of states it is easy to efficiently send messages from one set to the other by the means of an adjacency matrix.
This procedure is repeated for $k$ iterations, and ~\citet{scene-graph-iterative} found that $k=2$ gives optimal results.

\paragraph{Relation proposal network (RelPN)}
\citet{scene-graph-iterative} first proposed to build a fully connected graph connecting all object pairs and scoring relations between all possible object pairs; however, this model is expensive and grows exponentially with the number of objects.
\citet{graph-rcnn} introduced a \gls{relpn}, which works similarly to an object detection \gls{rpn} but that selectively proposes relations between pairs of objects.
In all our experiments, we use the Iterative Message Passing model trained using a \gls{relpn}.

\section{Conditional Graph Attention (C-GAT)}\label{sec:cgat}
Standard graph neural architectures encode information about neighbour nodes $\mathcal{N}(v)$ into representations of node $v \in \mathcal{V}$.
Therefore, these \glspl{gnn} are \textit{contextual} because they encode \textit{graph-internal} context.

We propose the {\bf \acrfull{cgat}} architecture, a novel extension for graph attention networks~\citep{graph_attention}.\footnote{This architecture is novel to the best of our knowledge.}
Our goal is to make these networks \textit{conditional} in addition to contextual.
By conditional we mean that a \gls{cgat} layer is conditioned on \textit{external} context, e.g. a vector representing knowledge that is not part of the original input graph.

Our motivation is that when using graph-based inputs such as a scene graph, a \gls{cgat} layer allows us to condition the message propagation between connected nodes in the graph on the current state of the model, e.g. on the decoder state in the captioning decoder in Figure~\ref{fig:cgat_model}.
Whereas a standard \gls{gat} layer contextually updates object hidden states, it cannot condition on context outside the scene graph.\footnote{This is generally true for standard \gls{gnn} architectures and not just \glspl{gat}.}
With a \gls{cgat} layer, we provide a mechanism for the model to learn to update object hidden states \textit{in the context of the current state of the decoder language model}, which we expect to lead to better contextual features.

\begin{figure}[t!]
    \centering
    \includegraphics[width=.9\linewidth]{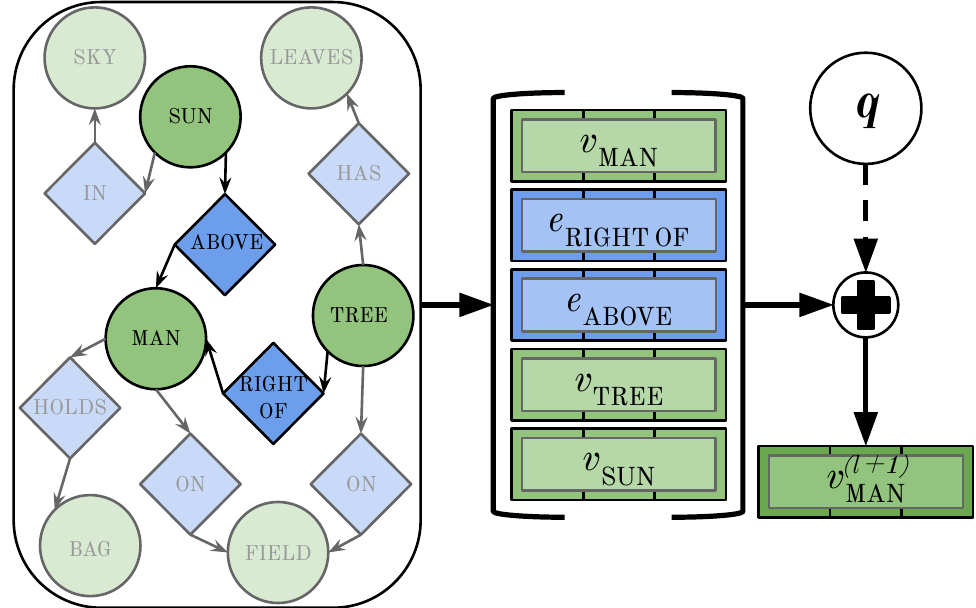}
    \caption{\gls{cgat} layer where we illustrate the update of vertex \texttt{MAN} by combining the features of all incoming relations (and objects) through an attention mechanism. The attention scores are conditioned on the external query vector $\bm{q}$.}
    \label{fig:cgat}
\end{figure}

In Figure~\ref{fig:cgat}, a \gls{cgat} layer is applied to an input scene graph $\mathcal{G}$ and conditioned on a query vector $\bm{q}$, i.e. the decoder state. We illustrate the update of vertex $v_\text{man} \in \mathcal{V}$ using features from its neighbourhood $\mathcal{N}(v_\text{man})$. The self-relation is assumed to always be present and for readability is not shown.
Neighbour nodes' features are combined with an MLP attention mechanism~\citep{bahdanau2014neural} and scores are computed using query $\bm{q}$.

As described in Section~\ref{sec:scene_graphs}, neighbours of an object vertex $v_i \in \mathcal{V}$ in a scene graph only include relation vertices.
To include neighbour object features as well as relation features, we collect features for all $v_j \in \mathcal{N}_\text{obj}(v_i)$, defined as nodes accessible by all relation vertices $v_r$ such that $\{v_i \leftarrow v_r, v_r \leftarrow v_j\} \in \mathcal{E}$.
%

\section{Model Setup}\label{sec:method}
In this section, we first introduce image captioning models that do not explicitly use relation features (Section~\ref{sec:ic}) and contrast them with those that use explicit relation features (Section~\ref{sec:ric}).

\subsection{Baseline Image Captioning (IC)}\label{sec:ic}
\ifglsused{ic}{}{\glsunset{ic}}

\paragraph{Bottom-Up Top-Down (BUTD)} 
The \acrlong{butd}~\citep[\acrshort{butd};][]{bottomup_topdown}\ifglsused{butd}{}{\glsunset{butd}} model consists of a Faster R-CNN image encoder ~\citep{faster-rcnn} that computes object proposal features for an input image, and a 2-layer LSTM language model decoder with a MLP attention mechanism over the object features that generates a caption for the image~\citep{hochreiter1997lstm, bahdanau2014neural}.
We denote the set of object features $\mathbf{X}\in\mathbb{R}^{n\times d}$, where $n$ is the number of objects in the image and $d$ the features dimensionality.
The 2-layer LSTM is designed so that the first layer is used to compute an attention over the image features
and the second layer is used to generate the captions' tokens.
LSTM states at time step $t$ are denoted as $\bm{h}_1^{(t)}$ and $\bm{h}_2^{(t)}$ for layer 1 and 2, respectively.
The hidden state of LSTM$_1$ is used to derive an attention over image features:
\begin{align}
\bm{x}^{(t)} = \mathit{Att}(\mathbf{X}, \bm{h}_1^{(t)}), \label{eq:att}
\end{align}
where $\bm{x}^{(t)} \in \mathbb{R}^d$ is the output of the attention layer and denotes the image features used at time step $t$.
Update rules for each LSTM layer are defined by:
\begin{align}
\bm{h}_1^{(t)} &= \text{LSTM}_1([\bm{h}_2^{(t-1)};\bm{w}^{(t-1)};\bar{\mathbf{X}}], \quad \bm{h}_1^{(t-1)}),\label{eq:lstm1}\\
\bm{h}_2^{(t)} &= \text{LSTM}_2([\bm{h}_1^{(t)};\bm{x}^{(t)}], \quad \bm{h}_2^{(t-1)}),\label{eq:lstm2}
\end{align}
where $\bm{w}^{(t-1)}$ is the embedding of the previously generated word, and $\bar{\mathbf{X}} \in \mathbb{R}^d$ are the mean image features. 

Next word probabilities are computed using a softmax over the vocabulary and parameterised by a linear projection of the hidden state of LSTM$_2$: $p(w^{(t)}=k|\bm{w}^{(1):(t-1)}) \propto \exp(\bm{W} \bm{h}_2^{(t)})$.

\subsection{Relation-aware Image Captioning (RIC)}\label{sec:ric}
\ifglsused{ric}{}{\glsunset{ric}}
We now describe 
models that incorporate explicit relation information into image captioning by using scene graphs as additional inputs.

We use the pretrained Iterative Message Passing model with a relation proposal network~\citep{scene-graph-iterative,graph-rcnn} to obtain scene graph features for all images. 
Scene graph features for an image are denoted $\mathbf{Y} \in \mathbb{R}^{(o+r) \times k}$, where $o$ is the number of objects, $r$ the number of relations between objects, and $k$ is the object/relation feature dimensionality.

We follow~\citet{wang-etal-2019-role} who have found that only using scene graph features led to poor results compared to using Faster R-CNN features only.
Therefore, we propose to integrate scene graph features
$\mathbf{Y}$ and Faster R-CNN object features $\mathbf{X}$ by experimenting with 
\textit{(i)} using $\mathbf{Y}$ directly, applying a \gls{gat} layer on $\mathbf{Y}$, or applying a \gls{cgat} layer on $\mathbf{Y}$ prior to feeding scene graph features into the decoder, and
\textit{(ii)} using a \acrlong{fa} mechanism versus a \acrlong{ha} mechanism.
%

\paragraph{GAT over Scene Graphs}
We propose a model that encodes the scene graph features $\mathbf{Y}$ with a standard GAT layer prior to using them in LSTM$_2$ in the decoder.

\paragraph{C-GAT over Scene Graphs}
In this setup, we apply a \gls{cgat} layer on scene graph features $\mathbf{Y}$ using the current decoder state $\mathbf{h}^{(t)}_1$ from LSTM$_1$  as the external context, and use the output of the \gls{cgat} layer in LSTM$_2$ in the decoder.

\paragraph{Flat Attention}
The \acrfull{fa}\ifglsused{fa}{}{\glsunset{fa}} consists of two separate attention heads, one over scene graph features $\mathbf{Y}$ and the other over Faster R-CNN features $\mathbf{X}$.
We use two standard MLP attention mechanisms \citep{bahdanau2014neural}, each using the hidden state from LSTM$_1$ as the query:
\begin{align}
\bm{x}^{(t)} = \mathit{Att}_x(\mathbf{X}, \bm{h}_1^{(t)}),\notag\\
\bm{y}^{(t)} = \mathit{Att}_y(\mathbf{Y}, \bm{h}_1^{(t)}).\notag
\end{align}
Each LSTM layer is now defined as follows:
\begin{align}
\bm{h}_1^{(t)} &= \text{LSTM}_1([\bm{h}_2^{(t-1)};\bm{w}^{(t-1)};\bar{\mathbf{X}};\bar{\mathbf{Y}}],\notag \bm{h}_1^{(t-1)}),\\
\bm{h}_2^{(t)} &= \text{LSTM}_2([\bm{h}_1^{(t)};\bm{x}^{(t)};\bm{y}^{(t)}], \quad \bm{h}_2^{(t-1)}),\label{eq:lstm_flat}
\end{align}
where $\bm{x}^{(t)}$ and $\bm{y}^{(t)}$ are computed by the two attention heads $Att_x$ and $Att_y$, respectively, and $\bar{\mathbf{Y}}$ denote the mean scene graph features.

\paragraph{Hierarchical Attention}
In a \acrfull{ha}\ifglsused{ha}{}{\glsunset{ha}} mechanism the output of the first attention head is used as input to derive the attention of the second head.
We again have two sets of inputs, scene graph features $\mathbf{Y}$ and Faster R-CNN object features $\mathbf{X}$.
We experiment first using $\mathbf{Y}$ as input to the first head, and its output $\mathbf{y}^{(t)}$ as additional input to the second head:
\begin{align}
\bm{y}^{(t)} &= \mathit{Att}_y(\mathbf{Y}, \bm{h}_1^{(t)}),\notag\\
\bm{x}^{(t)} &= \mathit{Att}_x(\mathbf{X}, [\bm{h}_1^{(t)};\bm{y}^{(t)}]).\label{eq:ha-sg}
\end{align}
This setup is similar to the cascade attention from~\citet{wang-etal-2019-role}.
We also try using $\mathbf{X}$ as input to the first head, and the first head's output $\mathbf{x}^{(t)}$ as additional input to the second head: 
\begin{align}
\bm{x}^{(t)} &= \mathit{Att}_x(\mathbf{X}, \bm{h}_1^{(t)}).\notag\\
\bm{y}^{(t)} &= \mathit{Att}_y(\mathbf{Y}, [\bm{h}_1^{(t)};\bm{x}^{(t)}]).\label{eq:ha-im}
\end{align}

In both cases, the hidden states for LSTM$_1$ and LSTM$_2$ are computed as in Equation~\ref{eq:lstm_flat}.

\section{Experimental Setup and Results}\label{sec:results}
We compare our models with the following external baselines with no access to scene graphs:
(1) the adaptive attention 
model  {\bf Add-Att} which determines at each decoder time step how much of the visual features should be used \citep{Lu-2017-knowing-when-to-look};
(2) the Neural Baby Talk model {\bf NBT} generates a sentence with gaps and fills the gaps using detected object labels \citep{neural_baby_talk};
(3) and the {\bf \gls{butd}} model \citep{bottomup_topdown} described in Section~\ref{sec:ic}.

We also compare with the following baselines that use scene graphs:
(1) The ``Know more, say less'' model {\bf KMSL} extracts features for objects and relations based on the scene graph, which are passed through two attention heads and finally combined using a flat attention head
\citep{know_more_say_less}; and (2) the {\bf Cascade} model~\citep{wang-etal-2019-role} which is similar to our hierarchical attention model with a GAT layer, but that instead uses a relational graph convolutional network~\citep{ marcheggiani-etal-2017-simple}.

We do not discuss model variants/results that are trained with an additional reinforcement learning step \citep{Rennie_2017_self_critical,yangetal2019autoencoding} and only compare single model results, since training and performing inference with such models is very costly and orthogonal to our research questions.

Our proposed models are: 
\acrlong{fa} ({\bf \acrshort{fa}}), \acrlong{hasg} ({\bf \acrshort{hasg}}) following Equation~\ref{eq:ha-sg}, \acrlong{haim} ({\bf \acrshort{haim}}) following Equation~\ref{eq:ha-im}, \acrshort{hasg} with \acrlong{gat} ({\bf \acrshort{hasg}+\acrshort{gat}}), and \acrshort{hasg} with \acrlong{cgat} ({\bf \acrshort{hasg}+\acrshort{cgat}}).
We choose the last two variants to extend \gls{hasg} following the setup used by \citep{wang-etal-2019-role}.

We evaluate captions generated by different models by investigating their SPICE scores \cite{anderson2016spice}, i.e. an F1 based semantic captioning evaluation metric computed over scene graphs. 
It uses the semantic structure of the scene graph to compute scores over several dimensions (object, relation, attribute, colour, count, and size). 

We use the MSCOCO \textit{karpathy split}~\citep{coco,deep_visualsemantic_alignments} which has 5k images each in validation and test sets, and we use the remaining 113k images for training. We build a vocabulary based on all words in the train split that occur at least 5 times.
We use MSCOCO evaluation scripts~\citep{coco} and report BLEU4~\citep[B4;][]{bleu}, 
CIDEr~\citep[C;][]{cider}, ROUGE-L~\citep[R;][]{rouge}, and SPICE~\citep[S;][]{anderson2016spice}. 
See Appendix~\ref{app:training} for extra information on our implementation and training procedures.

\begin{table}[t]
    \centering
    \begin{tabular}{l@{\hspace{6pt}}HHHrHrrr}
    \toprule
     & B1 & B2 & B3 & B4 & M & C & R & S \\
    \cmidrule{2-9}
    Add-Att${*^\dag}$  & 74.2 & 58.0 & 43.9 & 33.2 & 26.6 & 108.5 & ---  & --- \\
    NBT$^{\dag}$   & 75.5 & ---  & ---  & 34.7 & 27.1 & 107.2 & ---  & 20.1\\
    BUTD$^\dag$     & 77.2 & ---  & ---  & 36.2 & 27.0 & 113.5 & 56.4 & 20.3\\
    BUTD            & 75.4 & 59.2 & 45.5 & 34.8 & 27.0 & 109.2 & 55.7 & 20.0\\
    \cmidrule{2-9}
    Cascade$^\dag$  & ---  & ---  & ---  & 34.1 & \bf 26.8 & 108.6 & \underline{55.9} & \bf 20.3\\
    KMSL$^\dag$     & 76.7 & 59.8 & 45.3 & 33.8 & 26.2 & \bf 110.3 & 54.9 & 19.8\\
    \acrshort{fa}              & 74.6 & 58.6 & 44.7 & 33.7 & 25.4 & 102.5 & 54.7 & 18.8\\
    \acrshort{haim}           & 76.0 & 60.2 & 46.5 & \bf 35.7 & 26.6 & \underline{109.9} & \underline{55.9} & \underline{19.9}\\
    \acrshort{hasg}           & 75.7 & 59.7 & 45.9 & 35.0 & 26.5 & 109.1 & 55.7 & 19.8\\
    + \acrshort{gat}           & 75.2 & 59.5 & 45.7 & 34.7 & 26.2 & 106.4 & 55.4 & 19.4\\
    + \acrshort{cgat}      & 75.9 & 60.2 & 46.4 & \underline{35.5} & 26.6 & \underline{109.9} & \bf 56.0 & 19.8\\
    \bottomrule
    \end{tabular}
    \caption{Results on the MSCOCO test set, with models selected on the validation set (\textit{karpathy} splits). Models in the upper section do not use scene graphs, while those in the bottom section do.
    All models are trained to convergence for a maximum of 50 epochs.
    Metrics reported are: BLEU-4 (B4),
    CIDEr (C), ROUGE-L (R), and SPICE (S).
    See Section~\ref{sec:results} for details on all models and acronyms. We bold-face the best and underscore the second-best scores per metric (models that use scene graph).
    $^*$ Model uses features from last convolutional layer in CNN, i.e. no Faster R-CNN features. 
    $^\dag$ Results reported in the authors' original papers.
    }
    \label{tab:results}
\end{table}

\subsection{Image Captioning without Relational Features}\label{sec:exp_ic}
%
Our re-implementation of the \gls{butd} baseline scores slightly worse compared to the results reported by \citet{bottomup_topdown}.
This difference can be attributed to
the Faster R-CNN features used, i.e. we always use 36 objects per image whereas \citet{bottomup_topdown} use a variable number of objects per image (i.e. 10 to 100),
and there are other smaller differences in their training procedure.
Since all our models use these settings, in further experiments we compare to our implementation of the BUTD baseline.

\subsection{Image Captioning with relational features}\label{sec:exp_ric}
We notice that the KMSL model by \citet{know_more_say_less} slightly outperforms the other models according to CIDEr, while it performs worse in all other metrics. \citet{know_more_say_less} found performance increases when restricting the number of relations and report scores using this restriction, whereas we decided to use the full set of relations to test the effect of scene graph quality (see Section~\ref{sec:sg_quality}). Furthermore, the features used in the KMSL model are not directly extracted from the \gls{sgg} model as is the case for the other models, but an additional architecture is used for computing stronger features.

\paragraph{Flat vs. Hierarchical attention}
According to Table~\ref{tab:results}, \gls{fa} performs worse not only compared to \gls{ha} models, but also compared to other baselines.

The \gls{ha} model using Faster R-CNN object features in the first head, i.e. \gls{haim} setup, performs better than using the scene graph features first, i.e. \gls{hasg} setup.
We hypothesise that this difference comes from the additional guidance from $\mathbf{x}^{(t)}$ helping with a better attention selection over possibly more noisy features present in $\mathbf{Y}$. 
%

\paragraph{Additional GNN updates}\label{sec:add_gnn}
%
Directly using a \gls{gat} layer over scene graph features negatively impacts model performance. Comparing these results to the related Cascade model from \citet{wang-etal-2019-role}, we hypothesise that the R-GCN architecture works better in this setting, although compared to other models it still has lower scores according to most metrics.
The reason may be that the Cascade model by \citet{wang-etal-2019-role} was undertrained or could have used better hyperparameters, as indicated by our BUTD baseline performing comparably or better than their strongest model.\footnote{Our BUTD baseline scores 109.2 CIDEr, whereas their best model achieves 108.6 CIDEr.}

Combining a \gls{cgat} layer on the decoder improves overall results according to most metrics, though by a small margin.
This suggests that using additional \glspl{gnn} in the context of image captioning have a positive effect.
Furthermore, graph features learned using \gls{cgat} always outperform standard \gls{gat}, which coincides with our intuition that taking the current decoder hidden context into consideration can improve graph features.

\subsection{SPICE breakdown}\label{sec:spice}
\begin{table}[tb]
    \centering
    \begin{tabular}{l@{\hspace{6pt}}rHHrH@{}H@{}rH@{}H@{}}
    \toprule
    & All &&& Obj &&& Rel &&\\
    \cmidrule{2-9}
    BUTD    & \bf 19.8 & 54.3 & 12.4 & \bf 36.0 & 69.9 & 24.9 & 5.2 & 14.7 & 3.2\\
    \cmidrule{2-9} 
    FA      & 18.5 & 53.5 & 11.4 & 34.7 & 69.8 & 23.8 & 5.0 & 14.9 & 3.1\\
    HA-IM   & \underline{19.5} & 54.9 & 12.1 & \underline{35.9} & 70.6 & 24.8 & 5.2 & 15.4 & 3.3\\
    HA-SG   & \underline{19.5} & 55.4 & 12.1 & \underline{35.9} & 70.7 & 24.8 & 5.3 & 15.5 & 3.3\\
    + GAT   & 19.2 & 54.5 & 11.9 & 35.5 & 70.6 & 24.4 & \bf \underline{5.6} & 16.6 & 3.5\\
    + C-GAT & 19.4 & 55.6 & 12.0 & 35.8 & 71.4 & 24.6 & 5.3 & 15.4 & 3.3\\
    \bottomrule
    \end{tabular}
    \caption{Breakdown of overall SPICE scores (All) into 
    object (Obj) and relation (Rel) F1 scores.
    See Section~\ref{sec:results} for details on all models and acronyms. We bold-face the best overall scores and underline the best scores obtained by our models.
    }
    \label{tab:spice}
\end{table}
In our analysis, in addition to the overall SPICE F1 score for an entire caption, we break it down into scores over objects and over relations.\footnote{The SPICE score also includes the components attribute, colour, count, and size, but we do not report them directly.} This allows us to investigate how models are better or worse on describing objects and relations independently.
These results, computed for the validation split, are shown in Table~\ref{tab:spice}. 

When we look at individual scores for objects and relations, we notice a small and consistent gain in relation F1 by using scene graphs independently of the attention architecture or other design choices, but also observe lower object F1 scores with respect to the \gls{butd} baseline.
When object and relation scores are combined into a single F1 measure, it results in worse overall scores suggesting that the small increase in the relation scores is not sufficient to have a positive impact on captioning insofar.
%

\subsection{Scene Graph Quality}\label{sec:sg_quality}
\begin{figure}[t]
    \centering
    \includegraphics[width=.9\linewidth]{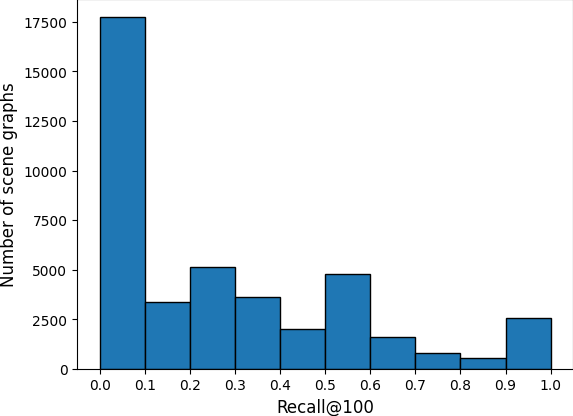}
    \caption{Distribution of the scores for the scene graphs in the validation split of the VG-COCO dataset.}
    \label{fig:sg_score_dist}
\end{figure}
\begin{table*}[t]
    \centering
    \subfloat[Full VG-COCO dataset\label{tab:vg-coco-full}]{
    \begin{tabular}{l@{\hspace{6pt}}rHHrHHrHHrHrr}
    \toprule
    & \multicolumn{9}{c}{SPICE}&\multicolumn{4}{c}{Captioning}\\
    \cmidrule{2-8} \cmidrule{11-14}
    & All & P & R & Obj & P & R & Rel & P & R & B4 & M & C & R\\
    \cmidrule{2-8} \cmidrule{11-14}
    BUTD    & \bf \underline{19.8} & 54.2 & 12.4 & \underline{36.0} & 70.1 & 24.9 & 5.0 & 14.3 & 3.1 & 35.4 & \bf 27.0 & 109.8 & \bf \underline{56.0}\\
    FA      & 18.5 & 53.3 & 11.4 & 34.9 & 70.1 & 23.9 & 4.8 & 14.6 & 3.0 & 33.2 & 25.4 & 103.6 & 54.8\\
    HA-IM   & 19.6 & 55.0 & 12.2 & \underline{36.0} & 71.1 & 24.8 & 5.2 & 15.2 & 3.3 & 35.0 & 26.6 & \underline{110.8} & 55.7\\
    HA-SG   & \bf \underline{19.8} & 56.3 & 12.2 & \bf 36.2 & 71.3 & 24.9 & \underline{5.5} & 16.3 & 3.4 & \bf \underline{35.7} & 26.7 & \bf 111.0 & \bf \underline{56.0} \\
    + GAT   & 19.4 & 55.0 & 12.0 & 35.5 & 71.0 & 24.4 & \bf 5.6 & 16.5 & 3.5 & 35.0 & 26.3 & 108.3 & 55.6\\
    + C-GAT & 19.5 & 55.8 & 12.1 & 35.8 & 71.6 & 24.5 & 5.2 & 15.1 & 3.2 & \bf \underline{35.7} & 26.6 & 110.4 & \bf \underline{56.0} \\
    \bottomrule
    \end{tabular}
    }
    \hspace{0.3em}
    \subfloat[Low VG-COCO dataset\label{tab:vg-coco-low}]{
    \begin{tabular}{HrHHrHHrHHrHrr}
    \toprule
    & \multicolumn{9}{c}{SPICE}&\multicolumn{4}{c}{Captioning}\\
    \cmidrule{2-8} \cmidrule{11-14}
    & All & P & R & Obj & P & R & Rel & P & R & B4 & M & C & R\\
    \cmidrule{2-8} \cmidrule{11-14} 
    BUTD    & \underline{19.5} & 53.2 & 12.2 & 35.6 & 68.9 & 24.8 & 4.7 & 13.3 & 2.9 & 34.0 & \bf 26.6 & 109.2 & 55.2\\
    FA      & 18.2 & 52.9 & 11.2 & 34.9 & 69.9 & 24.0 & 4.4 & 13.4 & 2.8 & 32.4 & 25.2 & 102.7 & 54.3\\
    HA-IM   & \underline{19.5} & 54.6 & 12.1 & \bf \underline{36.0} & 70.8 & 24.9 & \underline{5.2} & 15.1 & 3.2 & 34.3 & 26.4 & \bf 111.7 & \underline{55.5} \\
    HA-SG   & \bf 19.6 & 56.0 & 12.1 & \bf \underline{36.0} & 71.1 & 24.8 & 5.0 & 15.4 & 3.1 & 34.6 & 26.3 & \underline{110.3} & 55.4\\
    + GAT   & \underline{19.5} & 55.5 & 12.0 & 35.7 & 71.2 & 24.6 & \bf 5.7 & 16.7 & 3.6 & \bf \underline{34.8} & 26.3 & 109.5 & 55.4\\
    + C-GAT & \underline{19.5} & 55.5 & 12.0 & 35.7 & 71.0 & 24.5 & 5.1 & 14.7 & 3.2 & \bf \underline{34.8} & 26.4 & 109.9 & \textbf{55.7}\\
    \bottomrule
    \end{tabular}
    }\\
    \subfloat[Average VG-COCO dataset\label{tab:vg-coco-avg}]{
    \begin{tabular}{l@{\hspace{6pt}}rHHrHHrHHrHrr}
    \toprule
    & \multicolumn{9}{c}{SPICE}&\multicolumn{4}{c}{Captioning}\\
    \cmidrule{2-8} \cmidrule{11-14}
    & All & P & R & Obj & P & R & Rel & P & R & B4 & M & C & R\\
    \cmidrule{2-8} \cmidrule{11-14} 
    \acrshort{butd}    & \bf 20.5 & 55.5 & 12.8 & \bf 37.0 & 72.0 & 25.6 & \underline{5.5} & 16.2 & 3.4 & \bf 38.4 & \bf 27.6 & \bf 117.5 & \bf 57.1\\
    \acrshort{fa}      & 18.8 & 54.0 & 11.6 & 35.0 & 70.8 & 24.0 & 5.3 & 16.4 & 3.3 & 34.6 & 25.5 & 111.1 & 55.1\\
    \acrshort{haim}   & \underline{19.8} & 54.8 & 12.3 & \underline{36.2} & 71.3 & 25.0 & 5.3 & 15.4 & 3.3 & 36.2 & 26.5 & 115.1 & 55.3\\
    \acrshort{hasg}   & 19.6 & 55.5 & 12.1 & 35.9 & 71.3 & 24.7 & 5.4 & 16.2 & 3.3 & 37.0 & 26.4 & 114.4 & 56.3\\
    + \acrshort{gat}   & 19.4 & 54.5 & 12.0 & 35.8 & 71.7 & 24.6 & \bf 5.7 & 16.7 & 3.5 & 36.2 & 26.3 & 112.7 & 56.0\\
    + \acrshort{cgat} & 19.5 & 55.8 & 12.0 & 36.0 & 72.0 & 24.7 & 5.1 & 15.3 & 3.2 & \underline{37.6} & 26.5 & \underline{116.4} & \underline{56.1}\\
    \bottomrule
    \end{tabular}
    }
    \hspace{0.3em}
    \subfloat[High VG-COCO dataset\label{tab:vg-coco-high}]{
    \begin{tabular}{HrHHrHHrHHrHrr}
    \toprule
    & \multicolumn{9}{c}{SPICE}&\multicolumn{4}{c}{Captioning}\\
    \cmidrule{2-8} \cmidrule{11-14}
    Model & All & P & R & Obj & P & R & Rel & P & R & B4 & M & C & R\\
    \cmidrule{2-8} \cmidrule{11-14} 
    \acrshort{butd}    & \bf \underline{20.9} & 56.1 & 13.0 & \underline{36.8} & 72.6 & 25.3 & 5.1 & 14.9 & 3.2 & \underline{37.2} & \bf 27.8 & 126.5 & 57.0\\
    \acrshort{fa}      & 19.8 & 55.8 & 12.2 & 35.3 & 70.3 & 24.2 & 5.6 & 16.6 & 3.4 & 35.9 & 26.2 & 117.4 & 56.6\\
    \acrshort{haim}   & 20.3 & 57.0 & 12.6 & 36.2 & 71.9 & 24.9 & 5.8 & 17.5 & 3.6 & 36.2 & 27.3 & 124.1 & 56.8\\
    \acrshort{hasg}   & \bf \underline{20.9} & 59.0 & 12.9 & \bf 37.1 & 72.2 & 25.6 & \bf \underline{6.0} & 18.0 & 3.7 & \bf 38.1 & \bf 27.8 & \bf 129.8 & \bf 57.6\\
    + \acrshort{gat}   & 19.7 & 56.2 & 12.2 & 35.3 & 71.1 & 24.1 & 5.9 & 18.4 & 3.7 & 36.2 & 27.2 & 123.6 & \underline{57.1} \\
    + \acrshort{cgat} & 20.8 & 58.7 & 12.9 & 36.6 & 74.1 & 25.0 & \bf \underline{6.0} & 17.1 & 3.8 & \underline{37.2} & \bf 27.8 & \underline{127.3} & 56.9\\
    \bottomrule
    \end{tabular}
    }
    \caption{SPICE breakdown and captioning metrics for images in VG-COCO validation split. Results for the full VG-COCO, and for subsets of images collected according to the quality of their corresponding predicted scene graphs: low, average, and high.
    See Sections~\ref{sec:results} for details on all models and acronyms.
    Metrics reported are: overall SPICE F1 score (All), object (Obj) and relation (Rel) F1 score components, BLEU-4 (B4), CIDEr (C), and ROUGE-L (R).
    We bold-face the best and underline the second-best overall scores per metric and per data subset.
    }
    \label{tab:sg-quality}
\end{table*}
Since scene graph features are generated with a pretrained SGG model, we expect them to introduce a considerable amount of noise into the model.
In this section, we investigate the effect that the quality of the scene graph has on the quality of captions.

\paragraph{VG-COCO}
In this set of experiments, we need images with \textit{both} captions and scene graph annotations.
Thus, we use a subset of MSCOCO which overlaps with Visual Genome~\citep{krishnavisualgenome}, using captions from the former and scene graphs from the latter. We refer to this dataset as VG-COCO, as similarly done by \citet{know_more_say_less}.
We compute scores for each scene graph predicted by the Iterative Message Passing model using the common \textit{SGDet recall@100} as defined by \citet{graph-rcnn}. 
SGDet recall@100 is computed by using the 100 highest scoring triplets among all triplets predicted by the model,\footnote{A triplet is an object-predicate-subject phrase.} and reporting the percentage of gold-standard triplets.
The distribution of scores across images (Figure~\ref{fig:sg_score_dist}) shows that most scene graphs have extremely low scores close to zero, thus containing a lot of noise.

We separate images in the VG-COCO validation set in three groups:
low (R $<33\%$), average ($33\%\leq$ R $<67\%$), and high scoring graphs ($67\%\leq$ R), where R is SGDet recall@100.
For each set of images in each of these groups, we compute captioning metrics and also report a SPICE breakdown in Table~\ref{tab:sg-quality}. 

\paragraph{Effect of scene graph quality}
Due to the imbalance in scene graph quality, 
the \textit{low}, \textit{average}, and \textit{high} quality subsets have around 1000, 500, and 200 images, respectively.
By reporting results for the BUTD baseline, we show the performance a strong baseline obtains on the same set of images.

In Table~\ref{tab:vg-coco-low}, scores across all metrics are similar and only model FA performs clearly worse than others. 
Though the \gls{butd} baseline never performs best, it is often not more than a point behind the best performing model (except for CIDEr where it is 2.5 points lower compared to \acrshort{haim}). 

When comparing Table~\ref{tab:vg-coco-low} to Table~\ref{tab:vg-coco-avg}, we observe that all models tend to increase scores, and that \gls{butd} tends to perform best overall.
In Table~\ref{tab:vg-coco-high}, we see an increase in the difference between the baseline and our best models according to all metrics.
All these gains are very promising and suggest that when we have high quality scene graphs, we can expect a consistent positive transfer into image captioning models. However, the overall SPICE score is the highest for both \acrshort{butd} and \acrshort{hasg}, while \acrshort{butd} has lower scores for objects and relations F-measure.
That suggests that other components part of SPICE were worsened with the addition of scene graphs. Since this is not the focus of this paper, we did not investigate this further and leave that for future work.

Overall, these results show that indiscriminately using scene graphs from pretrained SGG models downstream on image captioning can be harmful because of the amount of noise present in these scene graphs.
However, when this noise is smaller and the scene graphs of higher quality, our findings together suggest that scene graphs can be useful in image captioning models.

\paragraph{Ground-truth graphs}
Finally, we also conduct a small-scale experiment using ground-truth scene graphs and evaluate how using these instead of predicted scene graphs at inference time impacts models, which can be found in Appendix~\ref{app:gt}.

\paragraph{Qualitative Results}
\begin{figure*}
    \centering
    \includegraphics[width=0.32\textwidth]{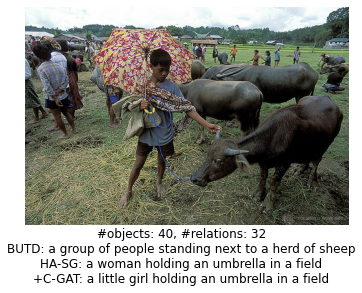}
    \includegraphics[width=0.32\textwidth]{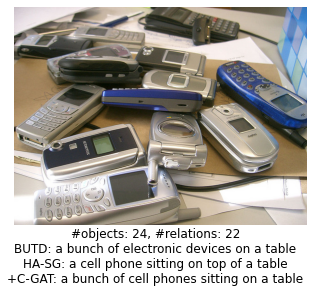}
    \includegraphics[width=0.32\textwidth]{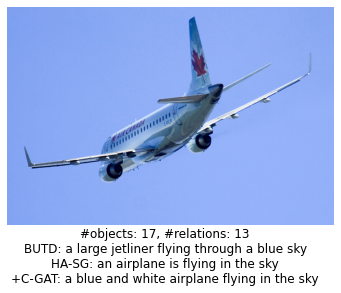}
    \caption{Images, captions and ground-truth number of objects and relations for \textbf{high} scoring scene graph.}
    \label{fig:qual_low}
\end{figure*}
\begin{figure*}
    \centering
    \includegraphics[width=0.32\textwidth]{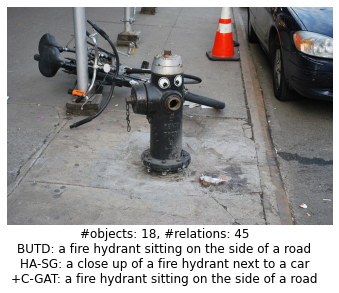}
    \includegraphics[width=0.32\textwidth]{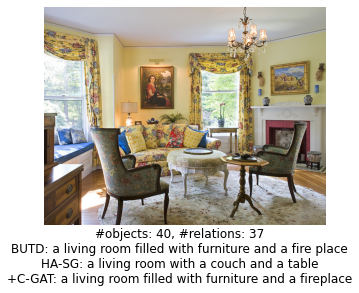}
    \includegraphics[width=0.32\textwidth]{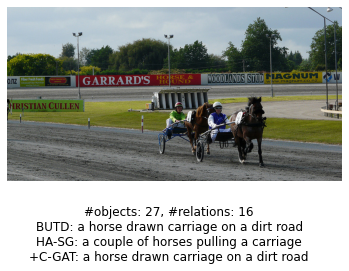}
    \caption{Images, captions and ground-truth number of objects and relations for \textbf{low} scoring scene graph.}
    \label{fig:qual_high}
\end{figure*}

Here, we try to determine if there is a clear difference in the difficulty in captioning images in \textit{low},  \textit{average}, and \textit{high} quality sets, which might help explain the result in Table~\ref{tab:sg-quality}. In Figures~\ref{fig:qual_low} and~\ref{fig:qual_high} we show some images for the low  and high scoring graphs, respectively. At a first glance, images from both sets appear equally cluttered with objects (i.e., which we hypothesise should correlate with the image being harder to describe). Furthermore, for both low and high scoring scene graphs, the average number of objects and relations is 23 and 22 respectively. However, we note that even scene graphs in the \textit{high} quality set often include tiny objects and details, e.g. the image in the right of Figure~\ref{fig:qual_low} shows a single aircraft, but there are 17 annotated objects describing components such as wings, windows, etc.

\section{Conclusions and Future Work}\label{sec:conclusions}
In this work, we investigate the impact scene graphs have on image captioning.
We introduced \acrfull{cgat} networks and applied it to image captioning, and report promising results (Table~\ref{tab:results}).
Overall, we found that improvements in captioning when using scene graphs generated with publicly available \gls{sgg} models are minor.
We observe a very small increase in the ability to describe relations as measured by relation SPICE F-scores, however, this is associated with models producing worse overall descriptions and producing lower object SPICE F-scores.

In an in-depth analysis, we found that the predicted scene graphs contain a large amount of noise which harms the captioning process. 
When this noise is reduced, large gains can be achieved across all image captioning metrics, e.g. 3.3 CIDEr points in the high VG-COCO split (Table~\ref{tab:vg-coco-high}).
This indicates that with further research and improved scene graph generation models, we will likely be able to observe consistent gains in image captioning and possibly other tasks by leveraging silver-standard scene graphs.

\paragraph{Future work}
In further research, we will conduct an in-depth analysis of our proposed conditional graph attention to determine
what tasks other than image captioning we can apply it to.
We envision using it for visual question-answering also with generated scene graphs, and on syntax-aware neural machine translation~\citep{bastings2017graph_nmt}, fake news detection~\citep{monti2019fake_news}, and question answering~\citep{zhang2018variational}.

In a focused qualitative analysis, we found that the scene graphs represent objects and relations in images sometimes with great detail. We plan to investigate
how to account for such highly detailed objects/relations in the context of image captioning.
Finally, we will look into a method to use predicted scene graphs selectively according to their estimated quality, possibly selecting the best graph between those generated by different SGG models.

\section*{Acknowledgments}
We would like to thank the COST Action CA18231 for funding a research visit to collaborate on this project.
This work is funded by the European Research Council (ERC) under the ERC Advanced Grant 788506.
IC has received funding from the European Union’s Horizon 2020 research and innovation program under the Marie Sk\l{}odowska-Curie grant agreement No 838188.

\bibliography{anthology,emnlp2020,bib1}
\bibliographystyle{acl_natbib}

\newpage\clearpage
\appendix

\section{Implementation Details}\label{app:training}
All our models are trained until convergence using early stopping with a patience of 20 epochs and a maximum of 50 epochs. 
We use the Adamax optimizer~\citep{kingma2014adam} with an initial learning rate of 0.002, which we decay with a factor of 0.8 after 8 epochs without improvements on the validation set.
Dropout regularisation with a probability of $50\%$ is applied on word embeddings and on the hidden state of the second LSTM layer $h_2^{(t)}$ before it is projected to compute the next word probabilities. 
We use a beam size of 5 during evaluation.
All hidden layers and embedding sizes are set to 1024. 
Models are all trained on a single 12GB NVIDIA GPU. 

We use a fixed number of $n=36$ objects extracted with our pretrained Faster R-CNN.
The number of objects and relations extracted with the pretrained Iterative Message Passing model varies according to the input image, i.e. a maximum of $o=100$ objects and of $r=2500$ relations.

\section{Using Ground-Truth Graphs}\label{app:gt}
\begin{table}[ht!]
\centering
\resizebox{\linewidth}{!}{%
\begin{tabular}{@{\hspace{3pt}}l@{\hspace{3pt}}rHHrHHr@{\hspace{4pt}}HH@{\hspace{4pt}}rHrr@{\hspace{3pt}}}
\toprule
& \multicolumn{9}{c@{\hspace{3pt}}}{SPICE}&\multicolumn{4}{c}{Captioning}\\
    \cmidrule{2-8} \cmidrule{11-14}
    & All & P & R & Obj & P & R & Rel & P & R & B4 & M & C & R\\
    
\cmidrule{2-8} \cmidrule{11-14} 
\acrshort{fa}      & 18.3 & 52.9 & 11.3 & 34.3 & 69.9 & 23.4 & \bf 5.3 & 15.4 & 3.4 & 30.5 & 24.0 & 95.8 & 53.1\\
\acrshort{haim}   & \bf 19.0 & 53.7 & 11.8 & \bf 35.2 & 70.3 & 24.2 & 4.8 & 14.1 & 3.0 & \underline{33.5} & 25.8 & \bf 106.0 & \bf 54.8 \\
\acrshort{hasg}   & 18.8 & 53.5 & 11.6 & 34.7 & 69.0 & 23.9 & 4.9 & 14.6 & 3.1 & 32.9 & 25.5 & \underline{104.5} & 54.3 \\
+ \acrshort{gat}   & 18.6 & 52.7 & 11.5 & 34.7 & 70.2 & 23.8 & \underline{5.2} & 15.4 & 3.3 & 32.5 & 25.0 & 100.9 & 53.9\\
+ \acrshort{cgat} & \underline{18.9} & 54.5 & 11.7 & \underline{34.8} & 71.2 & 23.7 & 5.1 & 14.8 & 3.2 & \bf 33.6 & \bf 25.6 & 104.3 & \underline{54.5} \\
\bottomrule
\end{tabular}%
}
\caption{Results for the full VG-COCO validation set using features extracted for ground-truth scene graphs. 
Models and acronyms are described in Sections~\ref{sec:results}.
Metrics reported are: the overall SPICE F1 score (All) and its object (Obj) and relation (Rel) F1 score components, BLEU-4 (B4), CIDEr (C), and ROUGE-L (R).
We bold-face the best and underline the second-best overall scores per metric.
}
\label{tab:gt}
\end{table}

In this small-scale experiment, we generate features for ground-truth scene graphs to determine if more a positive transfer can be achieved on image captioning models.
For the VG-COCO dataset, we take all the ground-truth object and relation boxes and pass these through the pretrained Iterative Message Passing (IMP) model, instead of the \gls{rpn} and \gls{relpn} proposed boxes. This is the same pretrained (IMP) model used in the other experiments. 

\citet{wang-etal-2019-role} also did a similar experiment, however, they also trained their models using features from gold-standard scene graphs, whereas we only use them to evaluate models previously trained on predicted scene graph features.
In Table~\ref{tab:gt} we show that when using ground-truth scene graphs results are worse than those obtained using predicted ones (Table~\ref{tab:vg-coco-full}).
One obvious explanation is the mismatch between training and testing data, with regards to quality and number of features. 
Models are trained on the predicted scene graphs, which have an average of 34 object and 48 relation features per image (probably noisy, as seen in Section~\ref{sec:sg_quality}), 
whereas ground-truth graphs have an average of 21 objects and 18 relations per image.

\end{document}